\documentclass{article}
\usepackage{spconf,amsmath,graphicx}
\usepackage{amsfonts} 
\usepackage{subcaption}
\usepackage{wrapfig}
\usepackage{url}

\DeclareMathOperator*{\argmax}{arg\,max}

\usepackage{longtable}
\usepackage{siunitx}
\usepackage{etoolbox}
\usepackage{multirow}
\renewcommand{\bfseries}{\fontseries{b}\selectfont}
\newrobustcmd{\B}{\bfseries} 
\title{AUTOMATIC LEARNING OF SUBWORD DEPENDENT MODEL SCALES}
\name{Felix Meyer$^{1}$, Wilfried Michel$^{1,2}$, Mohammad Zeineldeen$^{1,2}$, Ralf Schlüter$^{1,2}$, Hermann Ney$^{1,2}$}
  
  \address{$^{1}$Human Language Technology and Pattern Recognition, Computer Science Department,\\
RWTH Aachen University, 52074 Aachen, Germany \\
      $^{2}$AppTek GmbH, 52062 Aachen, Germany \\
      \small\texttt{felix.sebastian.meyer@rwth-aachen.de, \{michel,zeineldeen,schlueter,ney\}@cs.rwth-aachen.de}}

\begin{document}
%\ninept
%
\maketitle
\begin{abstract}
To improve the performance of state-of-the-art automatic speech recognition systems it is common practice to include external knowledge sources such as language models or prior corrections. This is usually done via log-linear model combination using separate scaling parameters for each model. Typically these parameters are manually optimized on some held-out data.
        
In this work we propose to optimize these scaling parameters via automatic differentiation and stochastic gradient decent similar to the neural network model parameters. We show on the LibriSpeech (LBS) and Switchboard (SWB) corpora that the model scales for a combination of attention-based encoder-decoder acoustic model and language model can be learned as effectively as with manual tuning. We further extend this approach to subword dependent model scales which could not be tuned manually which leads to 7\% improvement on LBS and 3\% on SWB. We also show that joint training of scales and model parameters is possible and gives additional 6\% improvement on LBS. 
\end{abstract}
\vspace{-7mm}
\section{Introduction}
\label{sec:intro}
\vspace{-2mm}
Attention-based encoder-decoder (AED) models \cite{rnn-aed, LAS} are investigated by many researchers in the field of automatic speech recognition (ASR) due to their simple modeling approach and end-to-end nature. It is, however, yet unclear how to best make use of unpaired text-only data. A common approach to increase the performance of AED models is the inclusion of an external language model (LM). These LMs are trained on text-only data and can therefore encode information from large text corpora that otherwise cannot be used directly.

The simplest solution to integrate an external LM is to combine the scores of acoustic model (AM) and LM via a log-linear combination. This approach is also called shallow fusion \cite{shallowfusion}. Previous investigations have shown that shallow fusion yields good results in most learning scenarios \cite{lm_integration_comparison}. In comparison with the usual setup, where AM and LM are trained separately and later combined, it is also possible to train both models jointly. In this case, parameters are often initialized with pretrained values. Investigations have shown that training the AM parameters while keeping an external LM fixed can yield good improvements \cite{early_stage_lm_integration}.

The benefit here is likely due to the suppression of an internal language model (ILM) in the decoder of the AED model. When AED models are trained, they learn an ILM from the transcriptions of the parallel training data \cite{transducer,ilm2}. Because of conflicts between internal and external LM it is worthwhile to subtract or suppress the ILM when including an additional external LM. An overview and a comparison of corresponding methods can be found in \cite{Zeineldeen.2021} and includes ILM estimation via context LMs.

In most of the mentioned approaches, different models are combined with the help of scale parameters that control the influence of each component. These scale parameters have to be tuned manually, which is usually done via grid search. In this work we introduce a method to learn these scale parameters via automatic differentiation \cite{backprop} and stochastic gradient decent \cite{adam}. This opens up the possibility to extend the log-linear combination to include more tunable parameters. We use this to investigate the use of individual scale parameters per output token, similar to \cite{hoffmeister09}.

\vspace{-2mm}
\section{Log-linear Model Combination}
\label{sec:models}
\vspace{-2mm}

\subsection{Subword Agnostic Scales}
\label{subseq:subword_agnostic_scales}
\vspace{-2mm}
When integrating an external LM into an ASR system, the definition of the posterior prediction probability changes. The probability distributions for the AM $p_{\mathrm{AM}}\left(w_1^N \mid x_1^T\right)$, and the LM $p_{\mathrm{LM}}(w_1^N)$ have to be combined. To receive a valid probability distribution, we have to perform a renormalization. We propose two ways to achieve this renormalization. The straight forward way is to perform it on a sentence level resulting in:
\begin{align}
    \widetilde{p}\left(w_1^N \mid x_1^T \right) &= \frac{p_{\mathrm{AM}}^{\alpha}(w_1^N | x_1^T) \cdot p_{\mathrm{LM}}^{\beta}(w_1^N)}{\sum_{\widetilde{w}_1^N}  p_{\mathrm{AM}}^{\alpha}(\widetilde{w}_1^N | x_1^T) \cdot p_{\mathrm{LM}}^{\beta}(\widetilde{w}_1^N)} \label{eq:p_sentence}
\end{align}
where $\alpha$ and $\beta$ $\in \mathbb{R}$ are scale parameters controlling the influence of AM and LM.  We note that in general, it is not possible to compute the above probability exactly, because of the sum over all possible sentences. In decoding, however, because the argmax is used to determine the best transcription and the denominator is independent of the argument, the normalization can be omitted. We note that in this case, only the ratio of the two parameters matters. Therefore, one of them can be fixed to $1$. This results in the following decision rule:
\begin{equation}
    \widetilde{w}_1^{\widetilde{N}} =  \argmax_{N,w_1^N} \left\lbrace  \log p_{\mathrm{AM}}\left(w_1^N \mid x_1^T\right) + \beta \log p_{\mathrm{LM}}(w_1^N) \right\rbrace \label{eq:shallow_fusion}
\end{equation}
In the literature, this way of combining AM and LM is commonly referred to as \emph{shallow fusion} \cite{shallowfusion, lm_integration_comparison}. 

The normalization can also be done on a per token basis:
\begin{align}
    \widehat{p}\!\left(w_1^N \mid x_1^T \right) \!&=\! \prod_{n=1}^N p_n\!\left(w_n \mid w_1^{n-1},x_1^T\right)\\
    &=\! \prod_{n=1}^N \!\frac{p_{\mathrm{AM},n}^{\alpha}(w_n | w_1^{n-1},x_1^T) \!\cdot\! p_{\mathrm{LM},n}^{\beta}(w_n | w_1^{n-1})}{\sum_{\widetilde{w}}  p_{\mathrm{AM},n}^{\alpha}(\widetilde{w} | w_1^{n-1},x_1^T) \!\cdot\! p_{\mathrm{LM},n}^{\beta}(\widetilde{w} | w_1^{n-1})} \label{eq:p_token}
\end{align}
We note that in this case both scale parameters matter, even when using the argmax. 
\vspace{-2mm}
\subsection{Subword Dependent Model Scales}
\label{subseq:subword_dependent_scales}
\vspace{-2mm}
In this work we extend the model scales by introducing individual scale parameters on a per BPE subword unit level. That is, each subword unit gets an individual AM and LM scale. We redefine $\alpha := \alpha_{w_1}, \dots, \alpha_{w_k} \in \mathbb{R}^k$ and $\beta := \beta_{w_1}, \dots, \beta_{w_k} \in \mathbb{R}^k$ where $w_1, \dots , w_k$ are all possible subword units and $k$ is the total number of subword units. The definition for the sentence level normalized prediction probability changes to
\begin{equation}
\begin{split}
    &\widetilde{p}\left(w_1^N \mid x_1^T \right) = \\ 
    & \quad \frac{\prod_{n=1}^N p_{\mathrm{AM},n}^{\alpha_{w_n}}(w_n| w_1^{n-1}, x_1^T) \cdot p_{\mathrm{LM},n}^{\beta_{w_n}}(w_n| w_1^{n-1})}{\sum_{\widetilde{w}_1^N} \prod_{n=1}^N p_{\mathrm{AM},n}^{\alpha_{\widetilde{w}_n}}(\widetilde{w}_n |\widetilde{w}_1^{n-1}, x_1^T) \cdot p_{\mathrm{LM},n}^{\beta_{\widetilde{w}_n}}(\widetilde{w}_n |\widetilde{w}_1^{n-1})} \label{eq:p_token_subword}
\end{split}
\end{equation}
and the token level normalized probability to
\begin{align}
    \widehat{p}(w_1^N | x_1^T ) &= \prod_{n=1}^N \frac{p_{\mathrm{AM},n}^{\alpha_{w_n}}(w_n | w_1^{n-1},x_1^T) \cdot p_{\mathrm{LM},n}^{\beta_{w_n}}(w_n | w_1^{n-1})}{\sum_{\widetilde{w}}  p_{\mathrm{AM},n}^{\alpha_{\widetilde{w}}}(\widetilde{w} | w_1^{n-1},x_1^T) \cdot p_{\mathrm{LM},n}^{\beta_{\widetilde{w}}}(\widetilde{w} | w_1^{n-1})}.
    \label{eq:p_sentence_subword}
\end{align}

\vspace{-5mm}
\section{Learning of Model Scales}
\label{sec:scales}
\vspace{-2mm}
Usually, the scale parameters from the decision rule of log-linear combination $\alpha$ and $\beta$ are tuned manually. This is commonly done via grid search, running the decoding process for different scale parameters. In this work, we propose to learn these parameters automatically. We use automatic differentiation \cite{backprop} of a training criterion $F$ and a variant of stochastic gradient descent \cite{adam} to find the optimal scale values similar to how other model parameters are optimized. 

\subsection{Training Criteria}
\label{subseq:train_crit}
\vspace{-2mm}
To train the scales we have to define a suitable training criterion. In analogy to the AM training we first use the cross entropy (CE) criterion. For simplicity we chose the per token renormalization from Equation \ref{eq:p_token}, which leads to
\begin{align}
    F_{\mathrm{CE}} 
    &= \log \widehat{p}\left(w_1^N \mid h_1^T \right)\\
    &= \sum_{n=1}^N \log \frac{p_{\mathrm{AM},n}^{\alpha}(w_n | w_1^{n-1},h_1^T) \cdot p_{\mathrm{LM},n}^{\beta}(w_n | w_1^{n-1})}{\sum_{\widetilde{w}}  p_{\mathrm{AM},n}^{\alpha}(\widetilde{w} | w_1^{n-1},h_1^T) \cdot p_{\mathrm{LM},n}^{\beta}(\widetilde{w} | w_1^{n-1})} \label{eq:ce_crit}.
\end{align}
This criterion does, however, not reflect the criterion that is used in the manual tuning process where the word error rate (WER) of the dev set is used directly. We therefore decided to also investigate the automatic learning of model scales with a minimum word error rate (minWER) training criterion similar to \cite{minwer,minwer2}. The training criterion uses the sentence level renormalization and is given by
\begin{equation}
    F_{\mathrm{minWER}} = \sum_{N,w_1^N}\widetilde{p}\left(w_1^N \mid h_1^T\right) \cdot \mathcal{A}\left(w_1^N,\widetilde{w}_1^{\widetilde{N}}\right)\label{minwer_crit}
\end{equation}
where $\widetilde{w}_1^{\widetilde{N}}$ is the correct transcription of the input audio and $\mathcal{A}\left(y,y'\right)$ is the accuracy of a token sequence $y$, treating $y'$ as the ground-truth. In practice it is not feasible to compute this sum exactly. Therefore, we use n-best lists to approximate the search space. The probability $p$ is being renormalized to this n-best list.

When training the subword dependent scale parameters introduced in Section \ref{subseq:subword_dependent_scales}, we use the same training criteria, replacing the single scales with the subword dependent scale parameters.

\begin{table*}[t]
    \centering
    \caption{Performance of learned subword agnostic and subword dependent scales trained on LibriSpeech using different training criteria and subsets of data, where $\alpha$ is the (average) AM scale and $\beta$ is the (average) LM scale}
    \vspace{2mm}
    \begin{tabular}{|c|c|c||S[table-format=1.2]|S[table-format=1.2]|S[table-format=1.2]||S|S|S|S|}
    \hline
     & \multicolumn{2}{c||}{scale training} & {\multirow{2}{*}{$\alpha$}} & {\multirow{2}{*}{$\beta$}} & {\multirow{2}{*}{$\frac{\beta}{\alpha}$}} & \multicolumn{2}{c|}{dev WER [\%]} & \multicolumn{2}{c|}{test WER [\%]} \\
     & criterion & set & & & & {clean} & {other} & {clean} & {other} \\
     \hline \hline
    baseline & - & - & {-} & {-} & {-} & 4.0 & 10.9 & 4.2 & 11.4 \\
    + LM & manual & dev-other & 2.77 & 1.00 & 0.36 & 2.9 & 8.3 & 3.2 & 9.0 \\
    \hline 
    \multirow{6}{*}{\shortstack{subword\\ agnostic\\ scales}} & \multirow{3}{*}{CE} & train & 1.24 & 0.26 & 0.21 & 3.0 & 8.7 & 3.2 & 9.3 \\
     & & dev-clean & 1.10 & 0.51 & 0.46 & 3.1 & 8.4 & 3.5 & 9.3 \\
     & & dev-other & 1.00 & 0.50 & 0.50 & 3.1 & 8.5 & 3.6 & 9.7 \\
    \cline{2-10}
     & \multirow{3}{*}{minWER} & train & 3.41 & 1.19 & 0.35 & 2.8 & 8.2 & 3.2 & 8.9 \\
     & & dev-clean & 3.14 & 0.89 & 0.28 & 2.9 & 8.3 & 3.2 & 9.1 \\
     & & dev-other & 2.82 & 1.46 & 0.52 & 3.1 & 8.3 & 3.5 & 9.2 \\
    \hline
    \multirow{6}{*}{\shortstack{subword\\ dependent\\ scales}} & \multirow{3}{*}{CE} & train & 1.49 & 0.63 & 0.45 & 2.8 & 8.0 & 3.1 & 8.5 \\
     & & dev-clean & 1.50 & 1.13 & 0.84 & 6.0 & 17.7 & 12.2 & 20.7 \\
     & & dev-other & 1.40 & 1.36 & 0.90 & 8.5 & 8.1 & 9.1 & 15.1 \\
    \cline{2-10}
     & \multirow{3}{*}{minWER} & train & 1.50 & 0.63 & 0.45 & 2.7 & 7.8 & \B 3.0 & \B 8.4 \\
     & & dev-clean & 1.50 & 0.64 & 0.45 & 2.5 & 7.9 & 3.1 & 8.5 \\
     & & dev-other & 1.50 & 0.64 & 0.45 & 2.7 & 7.4 & 3.1 & 8.5 \\
    \hline
\end{tabular}
    \label{tab:scales}
\end{table*}

\section{Experimental Setup}
\label{sec:experiments}
\vspace{-2mm}
For all of our experiments we use the RETURNN training framework \cite{returnn1, returnn2}. Configs are available online.\footnote{\url{https://github.com/rwth-i6/returnn-experiments/2022-scale-learning}}
We evaluate our methods on the LibriSpeech 960h \cite{librispeech} and Switchboard 300h \cite{Godfrey1992SWB} corpora.

Our acoustic models are attention-based encoder-decoder models with CNN+BLSTM encoder and a single layer LSTM decoder that predict subword units generated by byte-pair-encoding (BPE) (LBS:10025, SWB:534). The details of our LibriSpeech Model can be found in \cite{Zeyer.2019} and our Switchboard Model follows \cite{Zeineldeen.2021}.

As our language model for LibriSpeech we use a 4 layer LSTM based model with 140M parameters, the SWB LM is a 6 Layer transformer with 76M parameters, both trained on additional text data. 
In the experiments with joint training (Section \ref{subseq:joint_train}) we also use a single layer LSTM LM, which has the same size as the AM decoder. This LM was trained solely on the transcriptions of the LibriSpeech corpus.

\subsection{Training Procedure} 

As in the standard shallow fusion approach, we start by first training both AM and LM separately with the CE objective function. Afterwards, we combine them and initialize the scales randomly with mean 1.0 and small variance. Then we train only the introduced scale parameters while keeping the model parameters fixed. 

We investigate the joint training of the parameters by first training the scale parameters, AM and LM  as described above and then running a joint training phase. Here, we decided to still keep the LM parameters fixed, as prior investigations \cite{early_stage_lm_integration} have shown severe degradation when training the LM only on transcribed audio data.

\vspace{-5mm}
\section{Results}
\label{results}
\vspace{-2mm}
\subsection{Subword Agnostic Scale Training}
\vspace{-2mm}

We conduct experiments for subword agnostic scales, that is, one scale parameter for AM and the LM each. 
%In the manual training process we use grid search and determine a scale of $\alpha = 2.77$ while fixing $\beta = 1$ to work well for our models. In the automatic learning process 
We train the scales for 5 epochs on the train dataset or for 100 epochs on the dev sets with a conservative learning rate. It is likely possible to reduce the number of epochs.

For LibriSpeech, the results of the experiments for both presented training criteria and datasets are displayed in Table \ref{tab:scales}. Firstly, we observe that the scales learned with the cross entropy training criterion produce slightly worse results than the manually tuned ones. The results obtained from the scales learned with the minWER criterion reach the same performance as those obtained from manual tuning. 
For Switchboard, the results are presented in Table \ref{tab:swb_scales}. Here, the manual results can be found both by CE and minWER training.

In both cases, using the train set to estimate the scales seems to be more stable and to generalize better to the other test sets. This shows that our procedure can be used to automatically learn the scales for the shallow fusion method.

\vspace{-2mm}
\subsection{Subword Dependent Scale Training}
\vspace{-2mm}
We also learn subword dependent scales, that is one AM scale and one LM scale per subword unit, by training for 5 epochs on the train dataset or for 100 epochs on the dev sets. Additionally, for the minWER training criterion we initialize them with the scales obtained from the CE training step. 

The results on LibriSpeech for both the CE and the minWER criterion as well as the different training sets are displayed in Table \ref{tab:scales}. For the CE criterion we observe a clear improvement of 5.5 \% over the shallow fusion baseline on test-other when using the training dataset. When using the minWER criterion, the results are even better. For the training set we reach a relative improvement of 6.6 \% on the test-other dataset. When training on the dev sets, we see a much bigger improvement on the set we use for estimating the parameters but reduces generalization of the model.

The results for Switchboard are presented in Table \ref{tab:swb_scales}. We see that subword dependend scales achive the same performance as the manual baseline on Hub5, but show better generalization on RT03. No overfitting on Hub5'00 is observed when using it to tune the scales. Training with minWER criterion on the train set leads to the best results.

In both cases training on the dev sets is not stable enough to reach the baseline results.
% findings:
% converge early in general. for single train-min_wer, it took only 1 subepoch (split 6)
% generalizes much better on test sets
% does not work well with per label trained on dev similar to LibriSpeech

\vspace{-2mm}
\subsection{Joint Training}
\label{subseq:joint_train}
\vspace{-2mm}
For joint training we initialize the model with a pretrained AM, LM, and scale parameters as presented above. Afterwards, we continue fine tuning the AM parameters while keeping the LM parameters fixed. We run experiments for both fixed as well as trainable scale parameters. In the joint training phase we use the decoder sized LM mentioned in section \ref{sec:experiments} since it better matches the train set. After the joint training we replace this LM for the usual more powerful one and retune the scales again. We use the CE training criterion from Equation \eqref{eq:ce_crit} in all training steps.

We run the training on LibriSpeech for 5 epochs in each individual training step using the cross entropy criterion. The results of the experiments are displayed in Table \ref{tab:scales_joint}. We observe that the joint training phase yields clear improvements in WER of $13.8 \%$ relative on dev-other for subword agnostic scales and $6.3$ for subword dependent scales. We further observe that fixing the scale parameters during the joint training phase or continuing to tune them has a negligible effect. 

The results of subword agnostic and subword dependent scales after running the joint training phase are almost identical. This indicates that the advantage that is achieved by the subword dependent scales can also be learned by the AM.  

\tabcolsep=0.11cm
\begin{table}[tb]
    \centering
    \caption{Joint training results for subword agnostic and subword dependent scales traind on LibriSpeech with the CE criterion.}
    % maybe take out baseline values?
    \begin{tabular}{|c|c|c||c|S[table-format=2.1]|c|S[table-format=2.1]|}
    \hline
     & \multicolumn{2}{c||}{train}  & \multicolumn{2}{c|}{dev WER [\%]} & \multicolumn{2}{c|}{test WER [\%]} \\
     & AM & scales & clean & {other} & clean & {other} \\
     \hline \hline
%    baseline & no & - & 4.0 & 10.9 & 4.2 & 11.4 \\
%     + LM & no & - & 2.9 & 8.3 & 3.2 & 9.0 \\
%    \hline 
    \multirow{3}{*}{\shortstack[c]{subword\\agnostic}} & no & yes & 3.0 & 8.7 & 3.2 & 9.3 \\
    \cline{2-7}
     & \multirow{2}{*}{yes} & no & \B 2.6 & \B 7.5 & \B 2.8 & 8.0 \\
    \cline{3-7}
     & & yes & \B 2.6 & 7.6 & \B 2.8 & 8.2 \\
    \hline
     \multirow{3}{*}{\shortstack{subword\\dependent}} & no & yes & 2.8 & 8.0 & 3.1 & 8.5 \\
     \cline{2-7}
     & \multirow{2}{*}{yes} & no & \B 2.6 & \B 7.5 & \B 2.8 & 8.0 \\
     \cline{3-7}
     & & yes & \B 2.6 & \B 7.5 & \B 2.8 & \B 7.9 \\
    \hline
\end{tabular}
    \label{tab:scales_joint}
\end{table}

\begin{figure}[tbh]
    \centering
    \begin{subfigure}[b]{\linewidth}
    \includegraphics[width=0.975\linewidth]{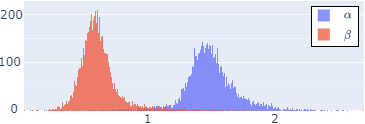}
    \caption{Distribution of AM scales ($\alpha$) and LM scales ($\beta$)}
    \label{fig:barplot}
    \end{subfigure}
    \begin{subfigure}[b]{0.475\linewidth}
    \includegraphics[width=\linewidth]{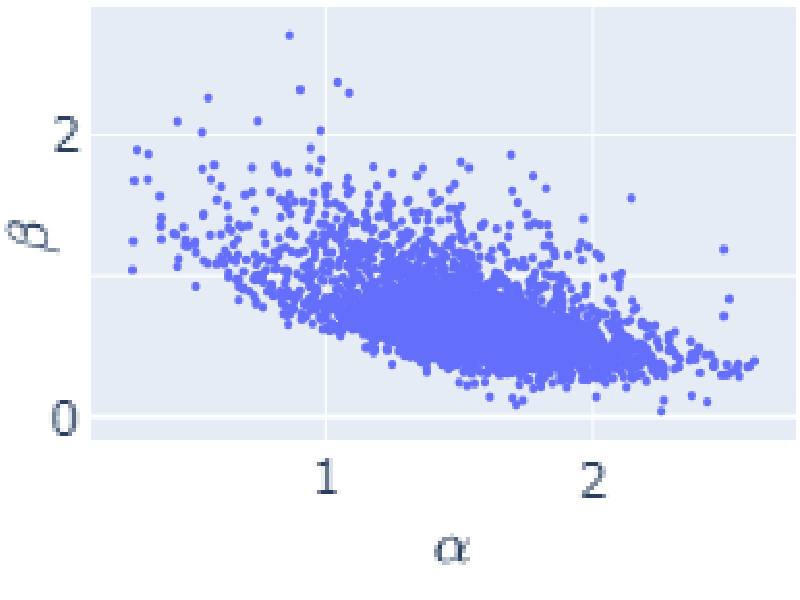}
    \caption{Correlation of $\alpha$ and $\beta$}
    \label{fig:correlation}
    \end{subfigure}
    \begin{subfigure}[b]{0.475\linewidth}
    \includegraphics[width=\linewidth]{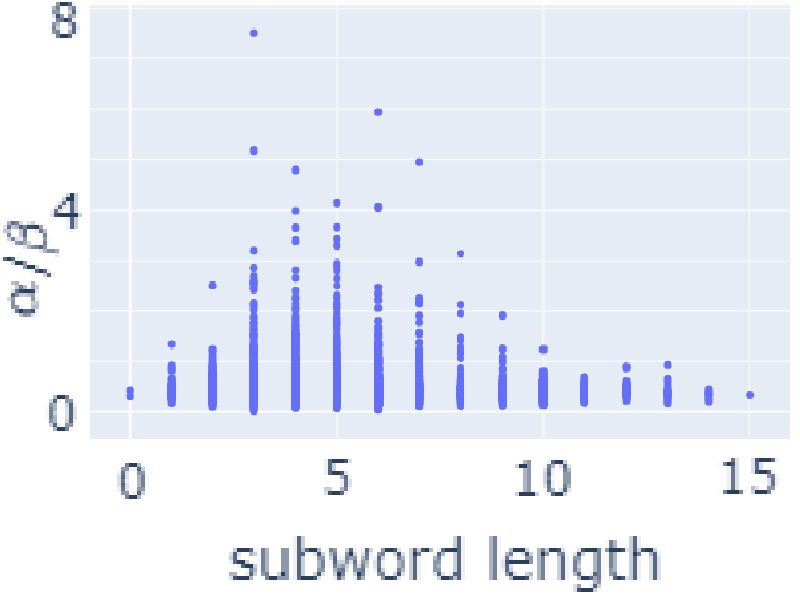}
    \caption{Length distribution}
    \label{fig:scale_length}
    \end{subfigure}
    \caption{Analysis of subword dependent AM scales $\alpha$ and LM scales $\beta$ trained on LibriSpeech 960h with minWER criterion}
\end{figure}

\subsection{Subword Dependent Scales Analysis}
\vspace{-2mm}
We analyse the scales found by training with the minWER training criterion on the train dataset of LibriSpeech. In Figure \ref{fig:barplot} we observe that the distribution of the scales follow a Gaussian distribution. When examining the correlation between AM and LM scales for each subword (cf. Figure \ref{fig:correlation}) we find a Pearson coefficient of $-0.52$. This implies a slight inversely proportional correlation. According to Figure \ref{fig:scale_length} the relative importance of AM and LM seems to be independent of the length of the BPE token in characters.

\begin{table}[t]
    \centering
    \caption{Performance of learned subword agnostic and subword dependent scales trained on Switchboard using different training criteria and subsets of data, where $\alpha$ is the (average) AM scale and $\beta$ is the (average) LM scale\vspace{-5mm}}
    \label{tab:swb_scales}
    \vspace{2mm}
    \begin{tabular}{|c|c|c||c|c|c|}
    \hline
     & \multicolumn{2}{c||}{scale training} & \multicolumn{3}{c|}{ WER [\%]} \\
     & criterion & set  & Hub5'00 & Hub5'01 & RT03 \\
     \hline \hline
    baseline & - & - & 12.3 & 11.9 & 14.3 \\
    + LM & manual & Hub5'00 & 12.1 & 11.7 & 14.1 \\
    \hline
    \hline
     \multirow{4}{*}{\shortstack{subword\\ agnostic\\ scales}} & \multirow{2}{*}{CE} & train & 12.1 & 11.6 & 13.8 \\
     & & Hub5'00 & 13.5 & 12.5 & 14.8 \\
    \cline{2-6}
     & \multirow{2}{*}{minWER} & train & 12.1 & 11.7 & 13.8 \\
     & & Hub5'00 & 12.1 & 11.6 & 13.8 \\
    \hline
     \multirow{4}{*}{\shortstack{subword\\ dependent\\ scales}} & \multirow{2}{*}{CE} & train & 12.1 & 11.7 & 13.9 \\ 
     & & Hub5'00 & 20.9 & 22.8 & 22.0 \\
     \cline{2-6}
     & \multirow{2}{*}{minWER} & train & \B 12.0 & \B 11.5 & \B 13.7 \\
     & & Hub5'00 & 14.0 & 14.5 & 16.5 \\
     \hline
     % No space left for the scales. Here is \alpha over \beta for reference
     % - 0.08 0.11 0.42 0.03 0.09 0.18 0.65 0.12 0.32
\end{tabular}
\end{table}

\section{Conclusion}
\label{sec:conclusion}
\vspace{-2mm}
In this work we proposed the automation of the tuning process for model scales used when combining different models via log-linear combination. 
%In particular, we successfully learned the scale parameters in a training setup with an attention-based encoder-decoder acoustic model and a neural language model. 
We conducted experiments with both cross entropy and  minimum word error rate training criterion on LibriSpeech and Switchboard and could recover the result of manual scale tuning. 
Training the scales on the whole training data showed better generalization of the scales to other test sets.

Additionally, we also extended the simple model scales by introducing individual scale parameters for each BPE subword unit. By doing this, we achieved a clear improvement of 6.6\% relative WER reduction on the LibriSpeech test-other dataset and 2.8\% improvement on the RT03 test set.

\section{Acknowledgements}
\label{ack}
\vspace{-2mm}
\begin{wrapfigure}[4]{l}{0.13\textwidth}
	\vspace{-6mm}
	\begin{center}
		\includegraphics[width=0.15\textwidth]{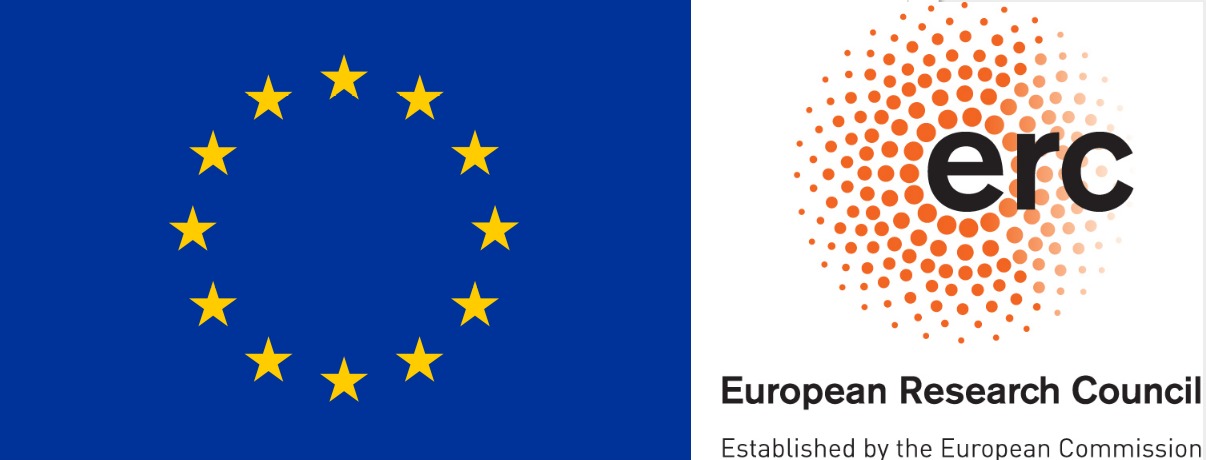} \\
	\end{center}
	\vspace{-6mm}
\end{wrapfigure}
\small 
This work has received funding from the European Research Council (ERC) under the European Union's Horizon 2020 research and innovation programme (grant agreement No 694537, project "SEQCLAS"). 
The work reflects only the authors' views and none of the funding parties is responsible for any use that may be made of the information it contains. 
%The GPU cluster used for the experiments was partially funded by Deutsche Forschungsgemeinschaft (DFG) Grant INST 222/1168-1. 

\normalsize

\vfill
\pagebreak

\bibliographystyle{IEEEbib}
\bibliography{strings,refs}

\end{document}